\documentclass{article}

\usepackage[square,numbers]{natbib}

\usepackage[preprint]{neurips_2020}

\usepackage[utf8]{inputenc} % allow utf-8 input
\usepackage[T1]{fontenc}    % use 8-bit T1 fonts
\usepackage{hyperref}       % hyperlinks
\usepackage{url}            % simple URL typesetting
\usepackage{booktabs}       % professional-quality tables
\usepackage{nicefrac}       % compact symbols for 1/2, etc.
\usepackage{microtype}      % microtypography

\usepackage{wrapfig}

\usepackage{caption}
\usepackage{subcaption}
\usepackage{tabularx}
\usepackage{todonotes}
\usepackage[group-separator={,}]{siunitx}

\usepackage{booktabs} % for toprule
\usepackage{multirow} % rotate text in tab 
\usepackage{hyperref}

\usepackage{amsmath,amsfonts,bm}
\usepackage{graphicx} 
\usepackage{xcolor}

\AtBeginDocument{%
  \providecommand\BibTeX{{%
    \normalfont B\kern-0.5em{\scshape i\kern-0.25em b}\kern-0.8em\TeX}}}

\title{The Transformer Network for the\\ Traveling Salesman Problem}

\author{%
  Xavier Bresson \\
  School of Computer Science and Engineering\\
  NTU, Singapore\\
  \texttt{xbresson@ntu.edu.sg} \\
  \And
  Thomas Laurent \\
  Department of Mathematics\\
  Loyola Marymount University\\
  \texttt{tlaurent@lmu.edu} \\
}

\begin{document}

\maketitle

\begin{abstract}
The Traveling Salesman Problem (TSP) is the most popular and most studied combinatorial problem, starting with von Neumann in 1951. It has driven the discovery of several optimization techniques such as cutting planes, branch-and-bound, local search, Lagrangian relaxation, and simulated annealing.
The last five years have seen the emergence of promising techniques where (graph) neural networks have been capable to learn new combinatorial algorithms. The main question is whether deep learning can learn better heuristics from data, i.e. replacing human-engineered heuristics? This is appealing because developing algorithms to tackle efficiently NP-hard problems may require years of research, and many industry problems are combinatorial by nature. 
In this work, we propose to adapt the recent successful Transformer architecture originally developed for natural language processing to the combinatorial TSP. Training is done by reinforcement learning, hence without TSP training solutions, and decoding uses beam search. We report improved performances over recent learned heuristics with an optimal gap of 0.004\% for TSP50 and 0.39\% for TSP100. 
\end{abstract}

\section{Traditional TSP Solvers}
The TSP was first formulated by William Hamilton in the 19th century. The problem states as follows; given a list of cities and the distances between each pair of cities, what is the shortest possible path that visits each city exactly once and returns to the origin city? TSP belongs to the class of routing problems which are used every day in industry such as warehouse, transportation, supply chain, hardware design, manufacturing, etc. TSP is an NP-hard problem with an exhaustive search of complexity O($n$!). TSP is also the most studied combinatorial problem. It has motivated the development of important optimization methods including Cutting Planes \cite{dantzig1954solution}, Branch-and-Bound \cite{bellman1962dynamic,held1962dynamic}, Local Search \cite{croes1958method}, Lagrangian Relaxation \cite{fisher1981lagrangian}, Simulated Annealing \cite{kirkpatrick1983optimization}. 

There exist two traditional approaches to tackle combinatorial problems; exact algorithms and approximate/heuristic algorithms. Exact algorithms are guaranteed to find optimal solutions, but they become intractable when $n$ grows. Approximate algorithms trade optimality for computational efficiency. They are problem-specific, often designed by iteratively applying a simple man-crafted rule, known as heuristic. Their complexity is polynomial and their quality depends on an approximate ratio that characterizes the worst/average-case error w.r.t the optimal solution.  

Exact algorithms for TSP are given by exhaustive search, Dynamic or Integer Programming. A Dynamic Programming algorithm was proposed for TSP in \cite{held1962dynamic} with O($n^2 2^n$) complexity, which becomes intractable for $n>40$.  A general purpose Integer Programming (IP) solver with Cutting Planes (CP) and Branch-and-Bound (BB) called Gurobi was introduced in \cite{gu2008gurobi}. Finally, a highly specialized linear IP+CP+BB, namely Concorde, was designed in \cite{applegate2006traveling}. Concorde is widely regarded as the fastest exact TSP solver, for large instances, currently in existence. \\

Several approximate/heuristic algorithms have been introduced. Christofides algorithm \cite{christofides1976worst} approximates TSP with Minimum Spanning Trees. The algorithm has a polynomial-time complexity with O($n^2 \log n$), and is guaranteed to find a solution within a factor $3/2$ of the optimal solution. Farthest/nearest/greedy insertion algorithms \cite{johnson1990local} have complexity O($n^2$), and farthest insertion (the best insertion in practice) has an approximation ratio of 2.43. Google OR-Tools \cite{googleOR} is a highly optimized program that solves TSP and a larger set of vehicle routing problems. This program applies different heuristics s.a. Simulated Annealing, Greedy Descent, Tabu Search, to navigate in the search space, and refines the solution by Local Search techniques. 2-Opt algorithm \cite{lin1965computer, johnson1995travelling} proposes an heuristic based on a move that replaces two edges to reduce the tour length. The complexity is O($n^2 m(n)$), where $n^2$ is the number of node pairs and $m(n)$ is the number of times all pairs must be tested to reach a local minimum (with worst-case being O($2^{n/2}$)). The approximation ratio is $4/\sqrt{n}$. Extension to 3-Opt move (replacing 3 edges) and more have been proposed in \cite{blazinskas2011combining}. Finally, LKH-3 algorithm \cite{helsgaun2017extension} introduces the best heuristic for solving TSP. It is an extension of the original LKH \cite{lin1973effective} and LKH-2 \cite{helsgaun2000effective} based on 2-Opt/3-Opt where edge candidates are estimated with a Minimum Spanning Tree \cite{helsgaun2000effective}. LKH-3 can tackle various TSP-type problems.

\section{Neural Network Solvers}
In the last decade, Deep learning (DL) has significantly improved Computer Vision, Natural Language Processing and Speech Recognition by replacing hand-crafted visual/text/speech features by features learned from data  \cite{lecun2015deep}. For combinatorial problems, the main question is whether DL can learn better heuristics from data than hand-crafted heuristics? This is attractive because developing algorithms to tackle efficiently NP-hard problems require years of research (TSP has been actively studied for seventy years). Besides, many industry problems are combinatorial. The last five years have seen the emergence of promising techniques where (graph) neural networks have been capable to learn new combinatorial algorithms with supervised or reinforcement learning. We briefly summarize this line of work below.

$\bullet$ HopfieldNets \cite{hopfield1985neural}: First Neural Network designed to solve (small) TSPs. \\
$\bullet$ PointerNets \cite{vinyals2015pointer}: A pioneer work using modern DL to tackle TSP and combinatorial optimization problems. This work combines recurrent networks to encode the cities and decode the sequence of nodes in the tour, with the attention mechanism. The network structure is similar to \cite{bahdanau2014neural}, which was applied to NLP with great success. The decoding is auto-regressive and the network parameters are learned by supervised learning with approximate TSP solutions. \\
$\bullet$ PointerNets+RL \cite{bello2016neural}: The authors improve \cite{vinyals2015pointer} with Reinforcement Learning (RL) which eliminates the requirement of generating TSP solutions as supervised training data. The tour length is used as reward. Two RL approaches are studied; a standard unbiased reinforce algorithm \cite{williams1992simple}, and an active search algorithm that can explore more candidates.   \\
$\bullet$ Order-invariant PointerNets+RL \cite{nazari2018reinforcement}: The original network \cite{vinyals2015pointer} is not invariant by permutations of the order of the input cities (which is important for NLP but not for TSP). This requires \cite{vinyals2015pointer} to randomly permute the input order to let the network learn this invariance. The work \cite{nazari2018reinforcement} solves this issue by making the encoder permutation-invariant. \\
$\bullet$ S2V-DQN \cite{dai2017learning}: This model is a graph network that takes a graph and a partial tour as input, and outputs a state-valued function Q to estimate the next node in the tour. Training is done by RL and memory replay \cite{mnih2013playing}, which allows intermediate rewards that encourage farthest node insertion heuristic. \\
$\bullet$ Quadratic Assignment Problem \cite{nowak2017note}: TSP can be formulated as a QAP, which is NP-hard and also hard to approximate. A graph network based on the powers of adjacency matrix of node distances is trained in supervised manner. The loss is the KL distance between the adjacency matrix of the ground truth cycle and its network prediction. A feasible tour is computed with beam search. \\
$\bullet$ Permutation-invariant Pooling Network \cite{kaempfer2018learning}: This work solves a variant of TSP with multiple salesmen. The network is trained by supervised learning and outputs a fractional solution, which is transformed into a feasible integer solution by beam search. The approach is non-autoregressive, i.e. single pass. \\
$\bullet$ Tranformer-encoder+2-Opt heuristic \cite{deudon2018learning}: The authors use a standard transformer to encode the cities and they decode sequentially with a query composed of the last three cities in the partial tour. The network is trained with Actor-Critic RL, and the solution is refined with a standard 2-Opt heuristic. \\
$\bullet$ Tranformer-encoder+Attention-decoder \cite{kool2018attention}: This work also uses a standard transformer to encode the cities and the decoding is sequential with a query composed of the first city, the last city in the partial tour and a global representation of all cities. Training is carried out with reinforce and a deterministic baseline. \\
$\bullet$ GraphConvNet \cite{joshi2019efficient}: This work learns a deep graph network by supervision to predict the probabilities of an edge to be in the TSP tour. A feasible tour is generated by beam search. The approach uses a single pass. \\
$\bullet$ 2-Opt Learning \cite{wu2019learning}: The authors design a transformer-based network to learn to select nodes for the 2-Opt heuristics (original 2-Opt may require O($2^{n/2}$) moves before stopping). Learning is performed by RL and actor-critic. \\
$\bullet$ GNNs with Monte Carlo Tree Search \cite{xing2020graph}: A recent work based on AlphaGo \cite{silver2016mastering} which augments a graph network with MCTS to improve the search exploration of tours by evaluating multiple next node candidates in the tour. This improves the search exploration of auto-regressive methods, which cannot go back once the selection of the nodes is made.

\section{Proposed Architecture}
\begin{figure*}[t]
\centering
\includegraphics[width=14cm]{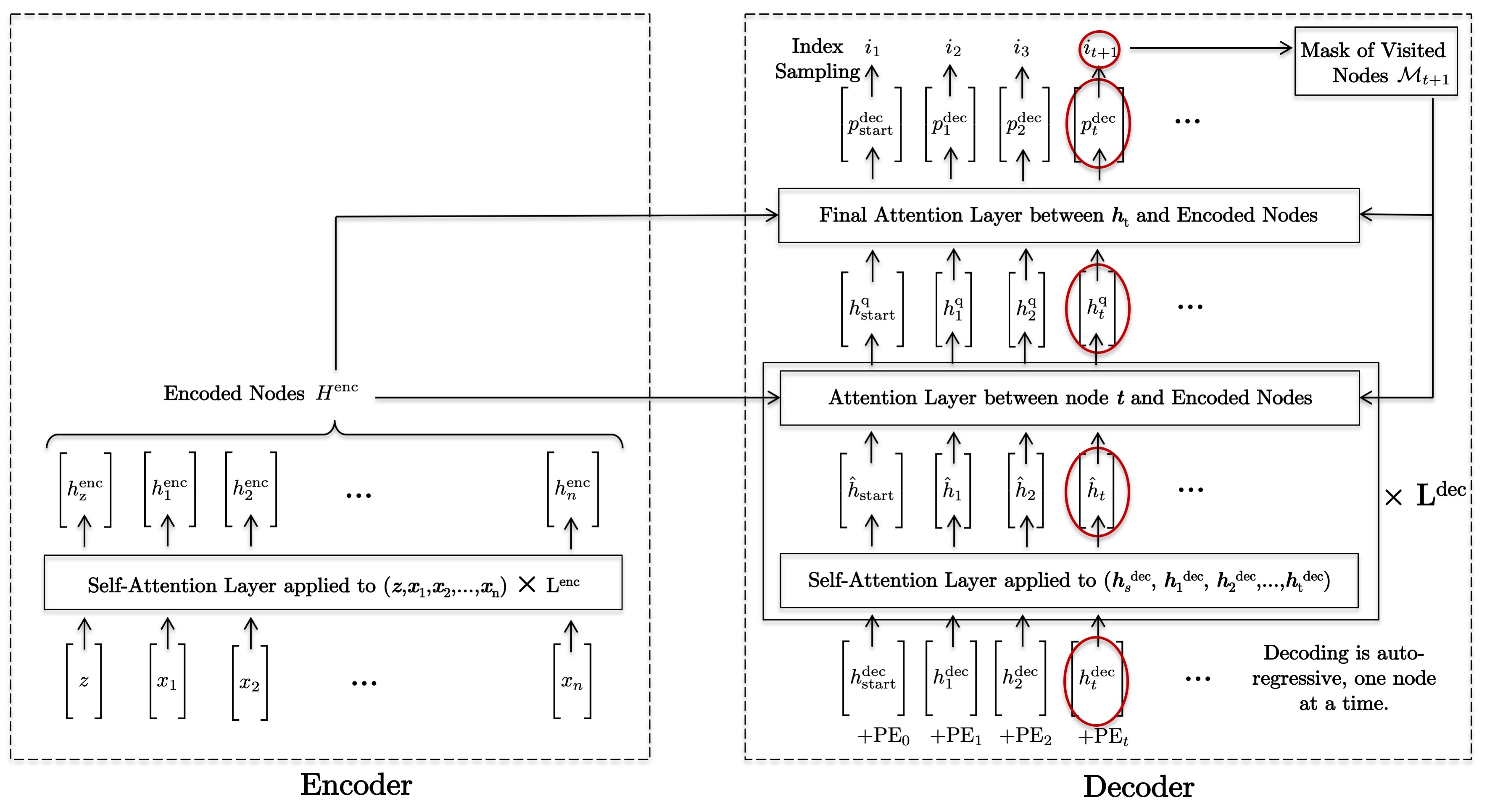}
\caption{Proposed TSP Transformer architecture.
}
\label{fig:net} 
\end{figure*}

We cast TSP as a ``translation'' problem where the source ``language'' is a set of 2D points and the target ``language'' is a tour (sequence of indices) with minimal length, and adapt the original Transformers \cite{vaswani2017attention} to solve this problem. We train by reinforcement learning, with the same setting as \cite{kool2018attention}. The reward is the tour length and the baseline is simply updated if the train network improves the baseline on a set of random TSPs. See Figure \ref{fig:net} for a description of the proposed architecture.

% FIGURE
% FIGURE

{\bf Encoder. } It is a standard Transformer encoder with multi-head attention and residual connection. The only difference is the use of batch normalization, instead of layer normalization. The memory/speed complexity is O($n^2$). Formally, the encoder equations are (when considering a single head for an easier description) 
\begin{eqnarray}
H^\textrm{enc}&=&H^{\ell=L^\textrm{enc}}\in\mathbb{R}^{(n+1)\times d},\\
\textrm{where}&& \nonumber \\
H^{\ell=0}&=&\textrm{Concat}(z,X)\in\mathbb{R}^{(n+1)\times 2}, z\in\mathbb{R}^{2}, X\in\mathbb{R}^{n\times 2},\\
H^{\ell+1} &=& \textrm{softmax}(\frac{Q^\ell {K^\ell}^T}{\sqrt{d}})V^\ell\in\mathbb{R}^{(n+1)\times d},\\
Q^\ell &=& H^\ell W_Q^\ell\in\mathbb{R}^{(n+1)\times d}, W_Q^\ell\in\mathbb{R}^{d\times d},\\
K^\ell &=& H^\ell W_K^\ell\in\mathbb{R}^{(n+1)\times d}, W_K^\ell\in\mathbb{R}^{d\times d},\\
V^\ell &=& H^\ell W_V^\ell\in\mathbb{R}^{(n+1)\times d}, W_V^\ell\in\mathbb{R}^{d\times d},\\
\end{eqnarray} 
where $z$ is a start token, initialized at random. See Figure \ref{fig:enc} for an illustration of the encoder.

% FIGURE
\vspace{0.25cm}
\begin{figure*}[h!]
\centering
\includegraphics[width=7cm]{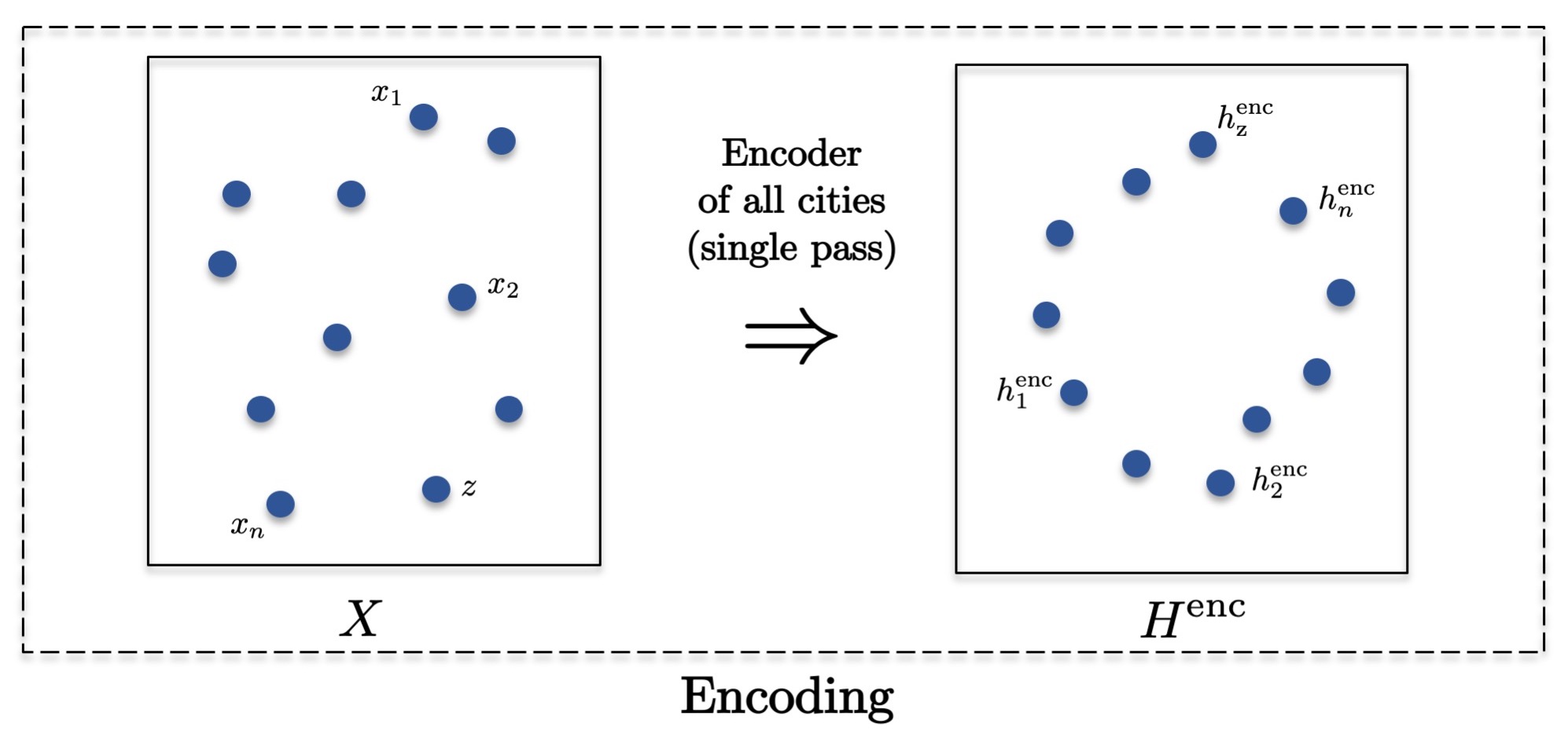}
\caption{Illustration of encoder.
}
\label{fig:enc} 
\end{figure*}
\vspace{0.25cm}
% FIGURE

\begin{figure*}[h!]
\centering
\includegraphics[width=14cm]{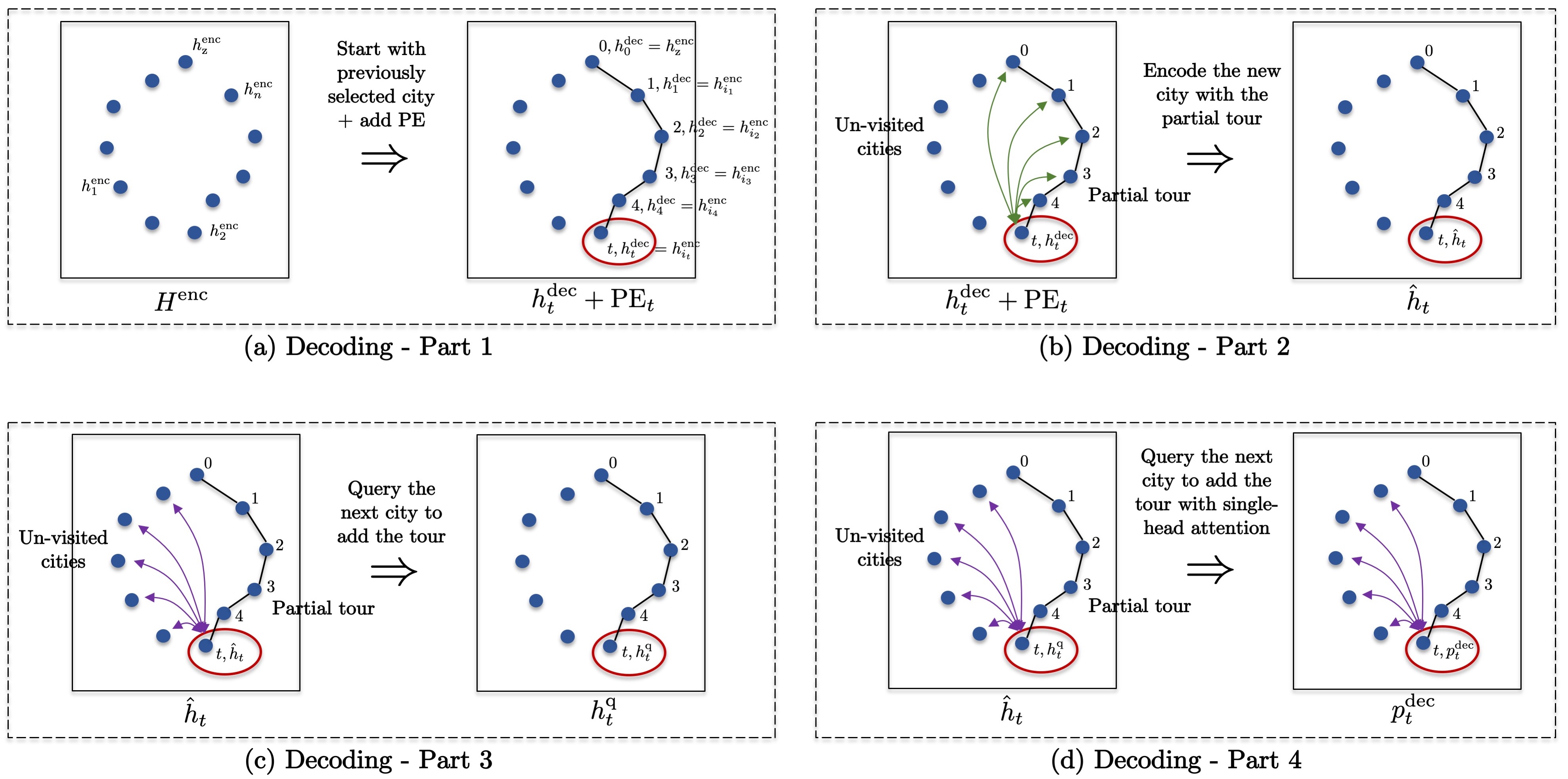}
\caption{Illustration of the four decoding steps.
}
\label{fig:dec} 
\end{figure*}

{\bf Decoder. } The decoding is auto-regressive, one city at a time. Suppose we have decoded the first $t$ cities in the tour, and we want to predict the next city. The decoding process is composed of four steps detailed below and illustrated on Figure \ref{fig:dec}.

% FIGURE
\vspace{0.25cm}
\vspace{0.25cm}
% FIGURE

{\bf Decoder – Part 1. } The decoding starts with the encoding of the previously selected $i_t$ city : 
\begin{eqnarray}
h_{t}^\textrm{dec}&=&h_{i_t}^\textrm{enc}+\textrm{PE}_t \in\mathbb{R}^{d},\\
h_{t=0}^\textrm{dec}&=&h_\textrm{start}^\textrm{dec}=z + \textrm{PE}_{t=0} \in\mathbb{R}^{d},
\end{eqnarray}
where $\textrm{PE}_t\in\mathbb{R}^{d}$ is the traditional positional encoding in \cite{vaswani2017attention} to order the nodes in the tour:
\begin{eqnarray}
\textrm{PE}_{t,i}=
\left\{
\begin{array}{lccc}
\sin (2\pi f_i t) \textrm{ if $i$ is even},  \\
\cos (2\pi f_i t) \textrm{ if $i$ is odd},\\
\end{array} 
\right.
\quad\textrm{with } f_i=\frac{10,000^\frac{d}{\lfloor 2i\rfloor}}{2\pi}. 
\end{eqnarray}

{\bf Decoder – Part 2. } This step prepares the query using self-attention over the partial tour. The self-attention layer is standard and uses multi-head attention, residual connection, and layer normalization. The memory/speed complexity is O($t$) at the decoding step $t$. The equations for this step are (when again considering a single head for an easier description) 
\begin{eqnarray}
\hat{h}_t^{\ell+1} &=& \textrm{softmax}(\frac{q^\ell {K^\ell}^T}{\sqrt{d}})V^\ell\in\mathbb{R}^{d},\ \ell=0,...,L^\textrm{dec}-1\\
q^\ell &=& \hat{h}_t^\ell \hat{W}_q^\ell\in\mathbb{R}^{d}, \hat{W}_q^\ell\in\mathbb{R}^{d\times d},\\
K^\ell &=& \hat{H}_{1,t}^\ell \hat{W}_K^{\ell}\in\mathbb{R}^{t\times d}, \hat{W}_K^\ell\in\mathbb{R}^{d\times d},\\
V^\ell &=& \hat{H}_{1,t}^\ell \hat{W}_V^{\ell}\in\mathbb{R}^{t\times d}, \hat{W}_V^\ell\in\mathbb{R}^{d\times d},\\
\hat{H}_{1,t}^\ell &=&[\hat{h}_1^\ell,..,\hat{h}_t^\ell], \ \hat{h}_t^\ell = 
\left\{
\begin{array}{ll}
h^\textrm{dec}_t \textrm{ if } \ell=0\\
h^{\textrm{q},\ell}_t \textrm{ if } \ell>0
\end{array}
\right..
\end{eqnarray}

{\bf Decoder – Part 3. } This stage queries the next possible city among the non-visited cities using a query-attention layer. Multi-head attention, residual connection, and layer normalization are used. The memory/speed complexity is O($n$) at each recursive step.
\begin{eqnarray}
h_t^{\textrm{q},\ell+1} &=& \textrm{softmax}(\frac{q^\ell {K^\ell}^T}{\sqrt{d}}\odot \mathcal{M}_t)V^\ell\in\mathbb{R}^{d},\ \ell=0,...,L^\textrm{dec}-1\\
q^\ell &=& \hat{h}_t^{\ell+1} \tilde{W}_q^\ell\in\mathbb{R}^{d}, \tilde{W}_q^\ell\in\mathbb{R}^{d\times d},\\
K^\ell &=& H^\textrm{enc} \tilde{W}_K^{\ell}\in\mathbb{R}^{t\times d}, \tilde{W}_K^\ell\in\mathbb{R}^{d\times d},\\
V^\ell &=& H^\textrm{enc} \tilde{W}_V^{\ell}\in\mathbb{R}^{t\times d}, \tilde{W}_V^\ell\in\mathbb{R}^{d\times d},\\ 
\end{eqnarray}
with $\mathcal{M}_t$ is the mask if the visited cities and $\odot$ is the Hadamard product.

{\bf Decoder – Part 4. } This is the final step that performs a final query using a single-head attention to get a distribution over the non-visited cities. Eventually, the next node $i_{t+1}$ is sampled from the distribution using Bernoulli during training and greedy (index with maximum probability) at inference time to evaluate the baseline. The memory/speed complexity is O($n$). The final equation is
\begin{eqnarray}
p_t^\textrm{dec}&=&\textrm{softmax}(C .\tanh (\frac{q K^T}{\sqrt{d}} \odot \mathcal{M}_t))\in\mathbb{R}^{n},\\
q &=& h_t^\textrm{q} \bar{W}_q \in\mathbb{R}^{d}, \bar{W}_q\in\mathbb{R}^{d\times d},\\
K &=& H^\textrm{enc} \bar{W}_K \in\mathbb{R}^{n\times d}, \bar{W}_K^\ell\in\mathbb{R}^{d\times d},\\ 
\end{eqnarray}
where $C=10$.

\section{Architecture Comparison}
Comparing Transformers for NLP (translation) vs. TSP (combinatorial optimization), the order of the input sequence is irrelevant for TSP but the order of the output sequence is coded with PEs for both TSP and NLP. TSP-Encoder benefits from Batch Normalization as we consider all cities during the encoding stage. TSP-Decoder works better with Layer Normalization since one vector is decoded at a time (auto-regressive decoding as in NLP). The TSP Transformer is learned by Reinforcement Learning, hence no TSP solutions/approximations required. Both transformers for NLP and TSP have quadratic complexity O($n^2 L$).

Comparing with the closed neural network models of \cite{kool2018attention} and \cite{deudon2018learning}, we use the same transformer-encoder (with BN) but our decoding architecture is different. We construct the query using all cities in the partial tour with a self-attention module. \cite{kool2018attention} use the first and last cities with a global representation of all cities as the query for the next city. \cite{deudon2018learning} define the query with the last three cities in the partial tour. Besides, our decoding process starts differently. We add a token city $z\in\mathbb{R}^2$. This city does not exist and aims at starting the decoding at the best possible location by querying all cities with a self-attention module. \cite{kool2018attention} starts the decoding with the mean representation of the encoding cities and a random token of the first and current cities. \cite{deudon2018learning} starts the decoding with a random token of the last three cities.

\section{Decoding Technique}
Given a set $X\in\mathbb{R}^{n\times 2}$ of 2-D cities, a tour is represented as an ordered sequence of city indices : $\textrm{seq}_n=\{i_1, i_2, …, i_n\}$ and TSP can be cast as a sequence optimization problem: 
\begin{eqnarray}
\max_{\textrm{seq}_n=\{i_1,...,i_n\}}\ P^\textrm{\tiny{TSP}}(\textrm{seq}_n|X)=P^\textrm{\tiny{TSP}}(i_1,...,i_n|X).
\end{eqnarray}

For auto-regressive decoding, i.e. selecting a city one at a time, $P^\textrm{\tiny{TSP}}$ can be factorized with the chain rule:
\begin{eqnarray}
P^\textrm{\tiny{TSP}}(i_1,...,i_n|X) = P(i_1|X)\hspace{0.05cm}\cdot\hspace{0.05cm} P(i_2|i_1,X) \hspace{0.05cm}\cdot\hspace{0.05cm} P(i_3|i_2,i_1,X) \cdot \ ...\ \cdot P(i_n|i_{n-1},i_{n-2},...,X).
\end{eqnarray}

Hence the decoding problem aims at finding the sequence $i_1, i_2, …, i_n$ that maximizes the objective:
\begin{eqnarray}
\max_{i_1,...,i_n}\ \Pi_{t=1}^n \ P(i_t|i_{t-1},i_{t-2},...i_1,X).
\end{eqnarray}

Finding exactly the optimal sequence by exhaustive search is intractable given the O($n!$) complexity, and approximations are necessary. The simplest approximate search is the greedy search; at each time step, the next city is selected with the highest probability:
\begin{eqnarray}
i_t = \arg\max_{i}\ P(i|i_{t-1},i_{t-2},...i_1,X)
\end{eqnarray}
The complexity is linear O($n$). 

Better sampling techniques such as beam search or Monte Carlo Tree Search (MTCS) are known to improve results over greedy search in NLP \cite{tillmann2003word} and TSP \cite{nowak2017note,kaempfer2018learning,kool2018attention,joshi2019efficient,wu2019learning,xing2020graph}. Their complexity is O($B n$), where $B$ is the number of beams or explored paths. Beam search \cite{lowerre1976harpy} is a breadth-first search (BFS) technique where the breath has a limited size $B$. the beam search decoding problem is as follows:
\begin{eqnarray}
\max_{\{i_1^b,...,i_n^b\}_{b=1}^B}\ \Pi_{b=1}^B \
P(i_1^b,...,i_n^b|X ) \ \textrm{ s.t. } \ \{i_1^b,...,i_n^b\} \not= \{i_1^{b'},...,i_n^{b'}\}, \ \forall b\not= b'
\end{eqnarray}

For $B=1$, the solution is given by greedy decoding. For $B>1$, the solution at $t$ is determined by considering all possible extensions of $B$ beams, and only keeping the Top-$B$ probabilities :
\begin{eqnarray}
\{i_1^b,...,i_t^b\}_{b=1}^B\ =\ \textrm{Top-B}\ \Big\{ \Pi_{k=1}^t  P(i_k^b|i_{k-1}^b,i_{k-2}^b,...,i_1^b,X )  \Big\}_{b=1}^{B.(n-t)},
\end{eqnarray}
or equivalently (for better numerical stabilities) :
\begin{eqnarray}
\{i_1^b,...,i_t^b\}_{b=1}^B\ =\ \textrm{Top-B}\ \Big\{ \sum_{k=1}^t \
\log P(i_k^b|i_{k-1}^b,i_{k-2}^b,...,i_1^b,X )  \Big\}_{b=1}^{B.(n-t)}.
\end{eqnarray}

\section{Numerical Experiments}
We compare the proposed architecture with existing methods in Table \ref{tab:results}. Our test set is composed of 10k TSP50 and TSP100. Concorde[1] run on Intel Xeon Gold 6132 CPU and the Transformers run on Nvidia 2080Ti GPU. Our code is available on GitHub \href{https://github.com/xbresson/TSP_Transformer}{https://github.com/xbresson/TSP\_Transformer}.

\begin{table}[t]
    \centering
        \vspace{3pt}
    \scalebox{0.75}{
    \begin{tabular}{ccc|cccc|cccc}
        \toprule
        && & \multicolumn{4}{c|}{\textbf{TSP50}} & \multicolumn{4}{c}{\textbf{TSP100}} \vspace{0.1cm}  \\
        && \textbf{Method} & \textbf{Obj} & \textbf{Gap} & \textbf{T Time} & \textbf{I Time} & \textbf{Obj} & \textbf{Gap} & \textbf{T Time} & \textbf{I Time} \\
        \midrule
        \parbox[t]{2mm}{\multirow{2}{*}{\rotatebox[origin=c]{90}{MIP}}} 
        && Concorde \cite{applegate2006traveling} & \textbf{5.689} & 0.00\% & 2m* & 0.05s & \textbf{7.765} & 0.00\% & 3m* & 0.22s \\
        && Gurobi \cite{gu2008gurobi} & - & 0.00\%* & 2m* & - & 7.765* & 0.00\%* & 17m* & - \\
        \midrule
        \parbox[t]{2mm}{\multirow{4}{*}{\rotatebox[origin=c]{90}{Heuristic}}} 
        && Nearest insertion & 7.00* & 22.94\%* & 0s* & - & 9.68* & 24.73\%* & 0s* & - \\
        && Farthest insertion \cite{johnson1990local} & 6.01* & 5.53\%* & 2s* & - & 8.35* & 7.59\%* & 7s* & -  \\
        && OR tools \cite{googleOR} & 5.80* & 1.83\%* & - & - & 7.99* & 2.90\%* & - & -  \\
        && LKH-3 \cite{helsgaun2017extension} & - & 0.00\%* & 5m* & - & \textbf{7.765}* & 0.00\%* & 21m* & -  \\
        \midrule
        \parbox[t]{2mm}{\multirow{9}{*}{\rotatebox[origin=c]{90}{\begin{tabular}{c} Neural Network\\Greedy Sampling \end{tabular} }}} 
        \hspace{0.2cm}
        && Vinyals et-al \cite{vinyals2015pointer} & 7.66* & 34.48\%* & - & -  & -  & -  & -  & - \\
        && Bello et-al \cite{bello2016neural} & 5.95* & 4.46\%* & - & -  & 8.30*  & 6.90\%*  & -  & - \\
        && Dai et-al \cite{dai2017learning} & 5.99* & 5.16\%* & - & -  & 8.31*  & 7.03\%*  & -  & - \\
        && Deudon et-al \cite{deudon2018learning} & 5.81* & 2.07\%* & - & -  & 8.85*  & 13.97\%*  & -  & - \\
        && Kool et-al \cite{kool2018attention} & 5.80* & 1.76\%* & 2s* & - & 8.12* & 4.53\%* & 6s* & - \\ 
        && Kool et-al \cite{kool2018attention} (our version) & - & - & - & - & 8.092 & 4.21\% & - & -  \\
        && Joshi et-al \cite{joshi2019efficient} & 5.87 & 3.10\% & 55s & - & 8.41 & 8.38\% & 6m & - \\
        && Our model & \textbf{5.707} & 0.31\% & 13.7s & 0.07s & \textbf{7.875} & 1.42\% & 4.6s & 0.12s \\
        \midrule
        \parbox[t]{2mm}{\multirow{11}{*}{\rotatebox[origin=c]{90}{\begin{tabular}{c} Neural Network \\ Advanced Sampling \end{tabular} }}}
        && Kool et-al \cite{kool2018attention} (B=1280) & 5.73* & 0.52\%* & 24m* & - & 7.94* & 2.26\%* & 1h* & - \\ 
        && Kool et-al \cite{kool2018attention} (B=5000) & 5.72* & 0.47\%* & 2h* & - & 7.93* & 2.18\%* & 5.5h* & - \\
        && Joshi et-al \cite{joshi2019efficient} (B=1280) & 5.70 & 0.01\% & 18m & - & 7.87 & 1.39\% & 40m & -  \\
        && Xing et-al \cite{xing2020graph} (B=1200) & - & 0.20\%* & - & 3.5s* & - & 1.04\%* & - & 27.6s*  \\
        && Wu et-al \cite{wu2019learning} (B=1000) & 5.74* & 0.83\%* & 16m* & - & 8.01* & 3.24\%* & 25m* & - \\
        && Wu et-al \cite{wu2019learning} (B=3000) & 5.71* & 0.34\%* & 45m* & - & 7.91* & 1.85\%* & 1.5h* & -  \\
        && Wu et-al \cite{wu2019learning} (B=5000) & 5.70* & 0.20\%* & 1.5h* & - & 7.87* & 1.42\%* & 2h* & -  \\
        && Our model (B=100) & 5.692 & 0.04\% & 2.3m & \textbf{0.09s} & 7.818 & 0.68\% & 4m & \textbf{0.16s}  \\
        && Our model (B=1000) & 5.690 & 0.01\% & 17.8m & 0.15s & 7.800 & 0.46\% & 35m & 0.27s  \\
        && Our model (B=2500) & \textbf{5.689} & 4e-3\% & 44.8m & 0.33s & \textbf{7.795} & 0.39\% & 1.5h & 0.62s  \\
 \bottomrule
    \end{tabular}
}
\vspace{0.25cm}
        \caption{Comparison with existing methods. Results with * are reported from other papers. 
T Time means total time for 10k TSP (in parallel).
I Time means inference time to run a single TSP (in serial).
}
\label{tab:results}
\end{table}

\begin{figure*}[h!]
\centering
\includegraphics[width=12cm]{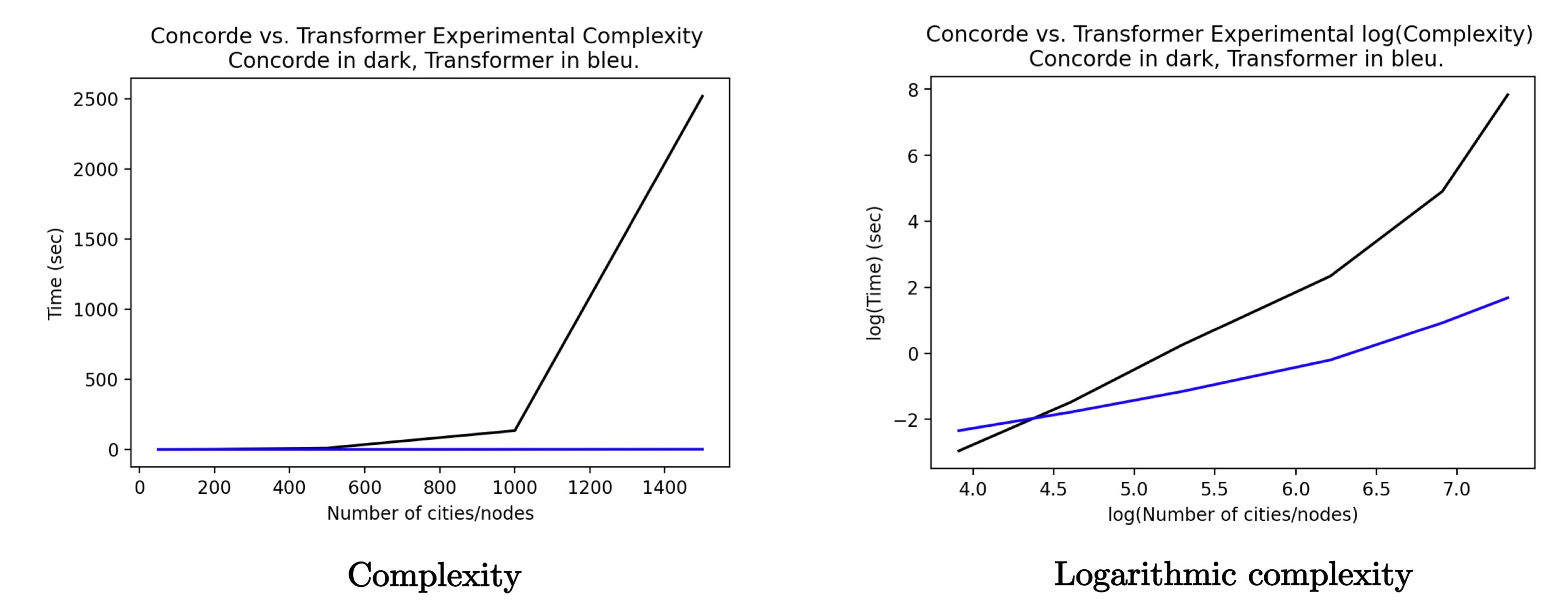}
\caption{Experimental complexity.
}
\label{fig:complexity} 
\end{figure*}

Experimental complexity for the inference time for a single TSP is presented on Figure \ref{fig:complexity}.

% FIGURE
% FIGURE

% FIGURE
\vspace{0.25cm}
\begin{figure*}[h!]
\centering
\includegraphics[width=14cm]{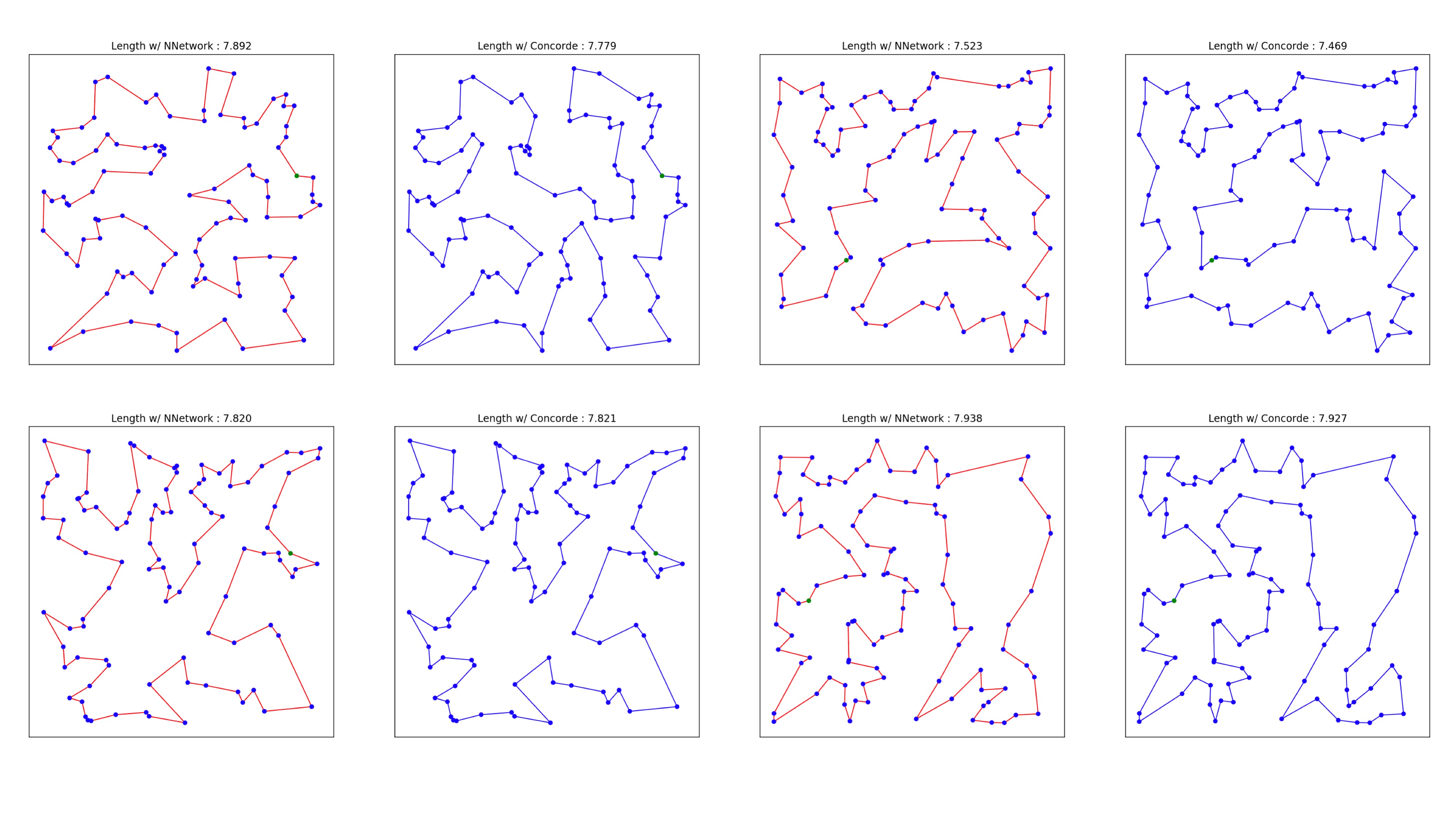}
\caption{Visualization of TSP100 instances.
}
\label{fig:enc} 
\end{figure*}
\vspace{0.25cm}
% FIGURE

\section{Discussion}
In this work, we essentially focused on the architecture. We observe that the Transformer architecture can be successful to solve the TSP Combinatorial Optimization problem, expanding the success of Transformer for NLP and CV. It also improves recent learned heuristics with an optimal gap of 0.004\% for TSP50 and 0.39\% for TSP100.

Further developments can be considered with better sampling techniques such as group beam-search \cite{vijayakumar2016diverse,meister2020best} or MCTS \cite{xing2020graph} which are known to improve results. Besides, the use of heuristics like 2-Opt to get intermediate rewards has also shown improvements \cite{wu2019learning} (the tour length as global reward requires to wait the end of the tour construction).

However, traditional solvers like Concorde/LKH-3 still outperform learning solvers in terms of performance and generalization, although neural network solvers offer faster inference time, O($n^2 L$) vs. O($n^{2.5} b(n)$), where O($b(n)$) is the number of branches to explore in BB. 

What’s next? The natural next step is to scale to larger TSP sizes for $n>100$ but it is challenging as GPU memories are limited, and Transformer architectures and auto-regressive decoding are in O($n^2$). We could consider “harder” TSP/routing problems where traditional solvers like Gurobi/LKH-3 can only provide weaker solutions or would take very long to solve. We could also work on “harder” combinatorial problems where traditional solvers s.a. Gurobi cannot be used. 

Another attractive research direction is to leverage learning techniques to improve traditional solvers. For example, traditional solvers leverage Branch-and-Bound technique \cite{bellman1962dynamic,held1962dynamic}. Selecting the variables to branch is critical for search efficiency, and relies on human-engineered heuristics s.a. Strong Branching \cite{achterberg2005branching} which is a high-quality but expensive branching rule. Recent works \cite{Gasse19,Nair20} have shown that neural networks can be successfully used to imitate expert heuristics and speed-up the BB computational time. Future work may focus on going beyond imitation of human-based heuristics, and learning novel heuristics for faster Branch-and-Bound technique.

\section{Conclusion}
The field of Combinatorial Optimization is pushing the limit of deep learning. Traditional solvers still provide better solutions than learning models. However, traditional solvers have been studied since the 1950s and the interest of applying deep learning to combinatorial optimization has just started. This topic of research will naturally expend in the coming years as combinatorial problems problems s.a. assignment, routing, planning, scheduling are used every day by companies. Novel software may also be developed that combine continuous, discrete optimization and learning techniques.

\section{Acknowledgement}
Xavier Bresson is supported by NRF Fellowship NRFF2017-10.

\bibliographystyle{ACM-Reference-Format}
\bibliography{tsp_transformer_bib}

\end{document}